%% file: main.tex
\documentclass[nohyperref]{article}

\usepackage{microtype}
\usepackage{graphicx}
\usepackage{subfigure}
\usepackage{booktabs} 

\usepackage{hyperref}

\usepackage[accepted]{icml2023}

\usepackage{amsmath}
\usepackage{amssymb}
\usepackage{mathtools}
\usepackage{amsthm}
\usepackage{makecell}
\usepackage{float}

\usepackage[capitalize,noabbrev]{cleveref}

\theoremstyle{plain}

\theoremstyle{definition}

\theoremstyle{remark}

\usepackage[textsize=tiny]{todonotes}

\icmltitlerunning{Tackling Shortcut Learning in Deep Neural Networks: An Iterative Approach with Interpretable Models}

\newcommand*\mystrut[1]{\vrule width0pt height0pt depth#1\relax}

\newcommand\ie {{\it i.e., }}

\newcommand\eg {{\it e.g., }}
\newcommand\etc{{\it etc.}}

\begin{document}

\twocolumn[
\icmltitle{Tackling Shortcut Learning in Deep Neural Networks: An Iterative Approach with Interpretable Models}



\begin{icmlauthorlist}
\icmlauthor{Shantanu Ghosh}{bu}
\icmlauthor{Ke Yu}{pitt}
\icmlauthor{Forough Arabshahi}{meta}
\icmlauthor{Kayhan Batmanghelich}{bu}
\end{icmlauthorlist}

\icmlaffiliation{bu}{Department of Electrical and Computer Engineering, Boston University, MA, USA}
\icmlaffiliation{pitt}{Intelligent Systems Program, University of Pittsburgh, PA, USA}
\icmlaffiliation{meta}{MetaAI, MenloPark, CA, USA}

\icmlcorrespondingauthor{Shantanu Ghosh}{shawn24@bu.edu}

\icmlkeywords{Machine Learning, ICML}

\vskip 0.3in
]



\printAffiliationsAndNotice{} 

\begin{abstract}
\label{sec:abstract}
\input{Sections/abstract_v1.tex}
\end{abstract}

\section{Introduction}
\label{sec:intro}

\input{Sections/intro_v1.tex}

\section{Method}
\label{sec:method}
\input{Sections/method_v1.tex}

\section{Experiments}
\label{sec:experiments}
\input{Sections/experiments_v1.tex}

\section{Results}
\textbf{MoIE does not compromise the performance of the original BB.}
\label{sec:quant}
\input{Sections/quant_v1.tex}


\textbf{Eliminating shortcuts.}
\label{sec:shortcuts}
\input{Sections/shortcut_v1.tex}

\section{Conclusion}
\label{sec:conclusion}
\input{Sections/conclusion_v1.tex}





\nocite{langley00}

\bibliography{example_paper}
\bibliographystyle{icml2023}

\newpage
\appendix
\onecolumn
\section{Appendix}
\subsection{Code}
\label{app:code}
Refer to the url \url{https://github.com/AI09-guy/ICML-Submission/} for the code.


Neuro-symbolic AI is an area of study that encompasses deep neural
networks with symbolic approaches to computing and AI to complement
the strengths and weaknesses of each, resulting in a robust AI capable
of reasoning and cognitive modeling~\cite{belle2020symbolic}.
Neuro-symbolic systems are hybrid models that leverage the robustness
of connectionist methods and the soundness of symbolic reasoning to
effectively integrate learning and reasoning
\cite{garcez2015neural,besold2017neural}. 




\subsection{Dataset}
\label{app:dataset}
\input{Appendix/dataset_v1.tex}

\subsection{Architectural details of symbolic experts and hyperparameters}
\label{app:g}

\input{Appendix/g_v1.tex}


\subsection{Expert driven explanations by MoIE}
\label{app:qual}
\input{Appendix/qual_v1.tex}

\subsection{Identification of harder samples by successive residuals}
\input{Appendix/harder_v1.tex}
\label{app:harder}



\end{document}

%% file: Sections/abstract_v1.tex
We use concept-based interpretable models to mitigate shortcut learning. Existing methods lack interpretability.
Beginning with a Blackbox, we iteratively \emph{carve out} a mixture of interpretable experts (MoIE) and a \emph{residual network}. Each expert explains a subset of data using First Order Logic (FOL). While explaining a sample, the FOL from biased BB-derived MoIE detects the shortcut effectively. Finetuning the BB with Metadata Normalization (MDN) eliminates the shortcut. The FOLs from the finetuned-BB-derived MoIE verify the elimination of the shortcut. Our experiments show that MoIE does not hurt the accuracy of the original BB and eliminates shortcuts effectively.

%% file: Sections/intro_v1.tex
Shortcuts pose a significant challenge to the generalizability of deep neural networks, denoted as Blackbox (BB), in real-world scenarios~\cite{geirhos2020shortcut, kaushik2019learning}. Referred to as spurious correlations, shortcuts indicate statistical associations between class labels and coincidental features that lack a meaningful causal connection. When trained on a dataset with shortcuts, a BB performs poorly when applied to test data without these shortcuts. This restricted generalization capability engenders a crucial concern, particularly in critical applications \eg medical diagnosis~\cite{bissoto2020debiasing}.

Various methods \eg invariant learning~\cite{arjovsky2020invariant}, correlation alignment~\cite{sun2016deep}, variance penalty~\cite{krueger2021outofdistribution}, gradient alignment~\cite{shi2021gradient}, instance reweighting~\cite{sagawa2019dro, liu2021just}, and data augmentation~\cite{Xu-2020, pmlr-v162-yao22b} have been employed to address the issue of shortcuts in Empirical Risk Minimization (ERM) models. However, they lack interpretability in 3 pivotal areas: 1) pinpointing the precise shortcut that the BB is aimed at, 2) clarifying the mechanism through which a particular shortcut is eradicated from the BB's representation, and 3) establishing a dependable technique to verify the elimination of the shortcut. The application of LIME~\cite{ribeiro2016should} and proxy-based interpretable models~\cite{rosenzweig2021patch} have been investigated to detect shortcuts in Explainable AI. However, they function within the pixel space rather than the human interpretable concept space~\cite{kim2017interpretability} and fail to address the issue of shortcut learning. This paper addresses this gap utilizing concept-based models.

Concept-based interpretable by design models~\cite{koh2020concept, zarlenga2022concept} use 1) a concept classifier to detect the presence/absence of concepts in an image, 2) an interpretable function (\eg linear regression or rule-based) to map the concepts to the final output.
However, these approaches utilize a single interpretable model to explain the whole dataset failing to encompass the diverse instance-specific explanations and exhibiting inferior performance than their BB counterparts.

\textbf{Our contributions.}
This paper proposes a novel method using the concept-based interpretable model to eliminate the shortcut learning problem. First we \emph{carve out} a mixture of interpretable models and a \emph{residual network} from a given BB. 
We hypothesize that a BB encodes multiple interpretable models, each pertinent to a unique data subset. 
As each interpretable model specializes over a subset of data, we refer to them as \emph{expert}.
Our design \emph{routes} the samples through the interpretable models to explain them with FOL. The remaining samples are routed through a residual network. On the residual, we repeat the method until all the experts explain the desired proportion of data. Next, we employ MDN~\cite{lu2021metadata}, a batch-level operation, to mitigate the impact of extraneous variables (metadata) on feature distributions. This approach effectively eliminates metadata effects during the training process.
Specifically, we deploy a 3-step procedure to mitigate the shortcuts: 
\begin{itemize}
\vspace{-1em}
\item[1)] The FOLs from biased BB detect the presence of the shortcut,
\vspace{-1em}
\item[2)] Assuming the detected shortcut as metadata, we use MDN layers to eliminate the shortcut by finetuning the BB,
\vspace{-1em}
\item[3)] The FOLs from the fine-tuned BB verify the elimination of the shortcut. 
\end{itemize}


%% file: Sections/method_v1.tex
\begin{figure}[t]
\begin{center}
\centerline{\includegraphics[width=\columnwidth]{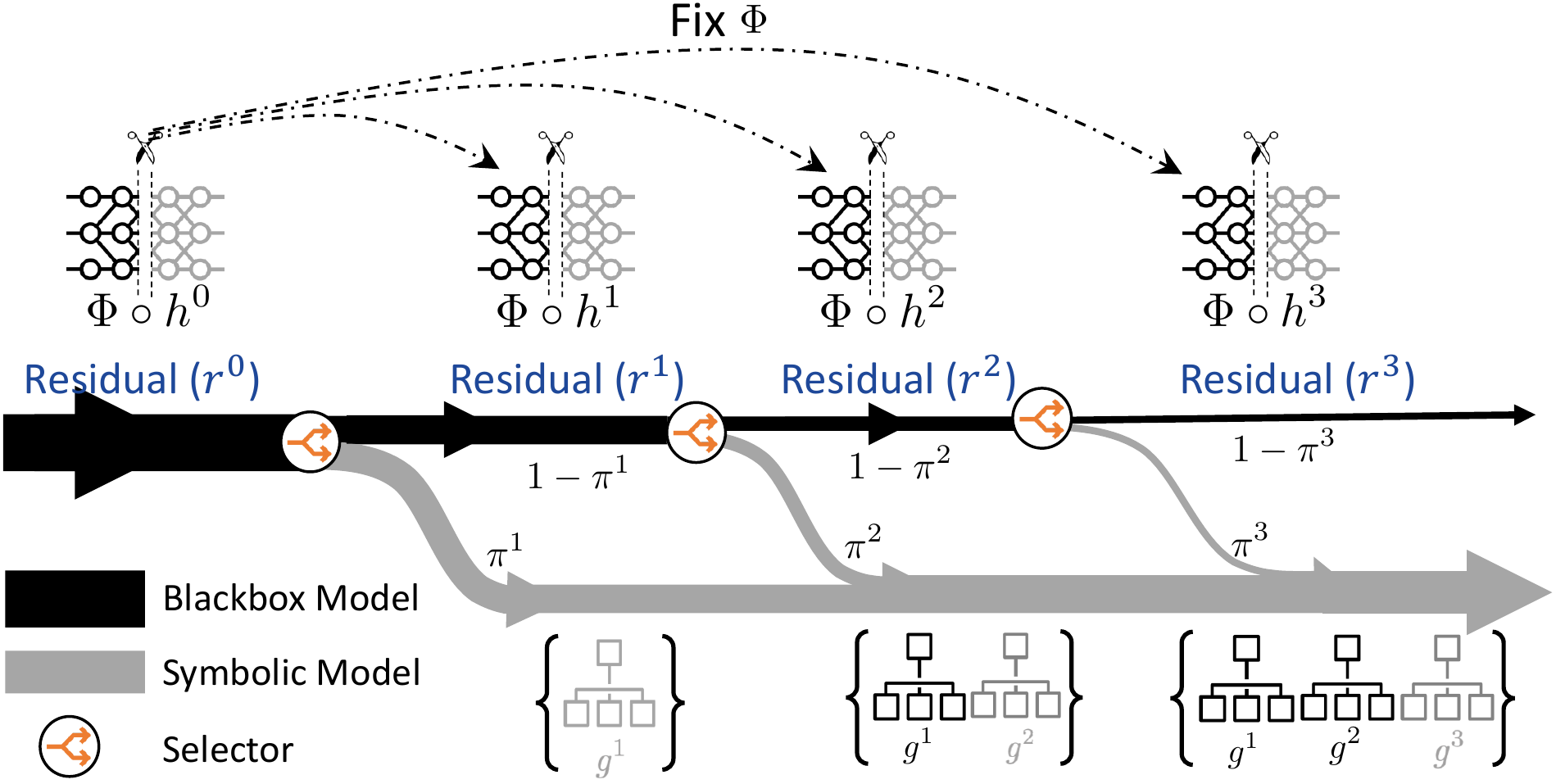}}
\caption{Schematic view of our method. Note that $f^k(.) = h^k(\Phi(.))$. At iteration $k$, the selector \emph{routes} each sample either towards the expert $g^k$ with probability $\pi^k(.)$ or the residual $r^k = f^{k-1} - g^k$ with probability $1-\pi^k(.)$. $g^k$ generates FOLs to explain the samples it covers. Note $\Phi$ is fixed across iterations.
}
\label{fig:Schematic} 
\end{center}
\vskip -0.2in
\end{figure}

\textbf{Notation:} 
Assume $f^0: \mathcal{X} \rightarrow \mathcal{Y}$ is a BB, on a dataset $\mathcal{X} \times\mathcal{Y} \times \mathcal{C}$, with $\mathcal{X}$, $\mathcal{Y}$, and $\mathcal{C}$ being the images, classes, and concepts, respectively; $f^0=h^0 \circ \Phi$, where $\Phi$ and  $h^0$ is the feature extractor and the classifier respectively. $f^0$ predicts $\mathcal{Y}$ from the input $\mathcal{X}$. Given a learnable projection~\cite{pmlr-v202-ghosh23c}, $t: \Phi \rightarrow \mathcal{C}$, our method learns three functions: (1) a set of selectors ($\pi: \mathcal{C}\rightarrow \{0, 1\}$) routing samples to an interpretable model or residual, (2) a set of interpretable models ($g: \mathcal{C} \rightarrow \mathcal{Y}$), and (3) the residuals. The interpretable models are called ``experts" since they specialize in a distinct subset of data defined by that iteration's coverage $\tau$ as shown in SelectiveNet~\cite{rabanser2022selective}. Fig.~\ref{fig:Schematic} illustrates our method.

\subsection{Distilling BB to the mixture of interpretable models}
\label{ns-optimization}
\textbf{The selectors:}
As the first step of our method, the selector $\pi^k$ \emph{routes} the $j^{th}$ sample through the interpretable model $g^k$ or residual $r^k$ with probability $\displaystyle \pi^k(\boldsymbol{c_j})$ and $\displaystyle 1 - \pi^k(\boldsymbol{c_j})$ respectively, where $k$ $\in [0,K]$, with $K$ being the number of iterations.
We define the empirical coverage of the $\displaystyle k^{th}$ iteration as $\vspace{-0.08pt} \zeta(\pi^k) = \frac{1}{m}\sum_{j = 1} ^ m \pi^k(\boldsymbol{c_j}) \vspace{-0.081pt}$, the empirical mean of the samples selected by the selector for the associated interpretable model $\displaystyle g^k$, with $\displaystyle m$ being the total number of samples in the training set. Thus, the entire selective risk is:
$
\mathcal{R}^k(\displaystyle \pi^k, \displaystyle g^k) = \frac{\frac{1}{m}\sum_{j=1}^m\mathcal{L}_{(g^k, \pi^k)}^k\big(\boldsymbol{x_j}, \boldsymbol{c_j}\big)}{\zeta(\pi^k)} ,
$
where $\mathcal{L}_{(g^k, \pi^k)}^k$ is the optimization loss used to learn $\displaystyle g^k$ and $\displaystyle \pi^k$ together, discussed in the next section. For a given coverage of $\displaystyle \tau^k \in (0, 1]$, we solve the following optimization problem:

\vskip -7.5pt
\begin{align}
\label{equ: optimization_g}
\theta_{s^k}^*, \theta_{g^k}^* = & \operatorname*{arg\,min}_{\theta_{s^k}, \theta_{g^k}} \mathcal{R}^k\Big(\pi^k(.; \theta_{s^k}), \displaystyle g^k(.; \theta_{g^k}) \Big) \nonumber \\ 
&\text{s.t.} ~~~ \zeta\big(\pi^k(.; \theta_{s^k})\big) \geq \tau^k,
\end{align}
\vskip 2pt

where $\theta_{s^k}^*, \theta_{g^k}^*$ are the optimal parameters at iteration $k$ for the selector $\pi^k$ and the interpretable model $g^k$ respectively. In this work, $\pi$s' of different iterations are neural networks with sigmoid activation. At inference time, the selector routes the $j^{th}$ sample with concept vector $\boldsymbol{c_j}$ to $\displaystyle g^k$ if and only if $\pi^k(\boldsymbol{c}_j)\geq 0.5$ for $k \in [0,K]$.

\begin{algorithm}[t]
   \caption{Applying MoIE to eliminate shortcuts}
   \label{alg:shortcut}
\begin{algorithmic}[1]
   \STATE {\bfseries Input:} $\mathcal{D}$ = \{$x_j$, $c_j$, $y_j$\}$_{j=1}^n$; biased BB $f^0 = h^0(\Phi(.))$; The total iterations K; Coverages $\tau_1, \dots ,\tau_K$. Freeze $\Phi$.
   \STATE Using~\cite{yuksekgonul2022post} learn the projection $t$ to predict the concept value.
   \STATE \textbf{Detection step.} Learn the experts in MoIE $\{g\}_{k=1}^K$ and extract the FOLs. The FOL contains shortcuts.
   \STATE \textbf{Elimination step.} Consider the detected shortcut concept in the ``Detection'' step as metadata and finetune BB ($f^0$) with MDN~\cite{lu2021metadata} to remove the role of that shortcut.
   \STATE Retrain $t$ with $\Phi$ of finetuned BB to get the concepts.
   \STATE \textbf{Verification step.} Learn MoIE $\{g\}_{k=1}^K$ again from retrained $t$ and recompute the FOLs. The final FOLs do not contain spurious concepts as they have been eliminated in the ``Elimination step''.
\end{algorithmic}
\end{algorithm}

\begin{table}[ht]
\caption{Datasets and BlackBoxes.}
\fontsize{5.2pt}{0.30cm}\selectfont
\label{tab:dataset}
\vskip 0.1in
\begin{center}
\begin{tabular}{lcc}
\toprule
DATASET & BB & \# EXPERTS  \\
\midrule
CUB-200~\cite{wah2011caltech}  & RESNET101~\cite{he2016deep}& 6 \\
CUB-200~\cite{wah2011caltech}  & VIT~\cite{wang2021feature}&  6 \\
AWA2~\cite{xian2018zero} & RESNET101~\cite{he2016deep} & 4\\
AWA2~\cite{xian2018zero} & VIT~\cite{wang2021feature} & 6\\
HAM1000~\cite{tschandl2018ham10000}    & INCEPTION~\cite{szegedy2015going}& 6\\
SIIM-ISIC~\cite{rotemberg2021patient}    & INCEPTION~\cite{szegedy2015going} & 6\\
EFFUSION IN MIMIC-CXR~\cite{12_johnsonmimic}    & DENSENET121~\cite{huang2017densely} & 3\\
\bottomrule
\end{tabular}
\end{center}
\vskip -0.1in
\end{table}

\textbf{The experts:}
For iteration $k$, the loss $\mathcal{L}_{(g^k, \pi^k)}^k$ distills the expert $g^k$ from $f^{k-1}$, BB of the previous iteration:
\begin{equation}
\label{equ: g_k}
\resizebox{0.47\textwidth}{!}{$
\mathcal{L}_{(g^k, \pi^k)}^k\big(\boldsymbol{x_j}, \boldsymbol{c_j}\big) = \underbrace{\mystrut{2.6ex}\ell\Big(f^{k - 1}(\boldsymbol{x_j}), g^k(\boldsymbol{c_j})\Big)\pi^k(c_j) }_{\substack{\text{trainable component} \\ \text{for current iteration $k$}}}\underbrace{\prod_{i=1} ^{k - 1}\big(1 - \pi^i(\boldsymbol{c_j})\big)}_{\substack{\text{fixed component trained} \\ \text{in the previous iterations}}},
$}
\end{equation}

where the term $\pi^k(\boldsymbol{c_j})\prod_{i=1} ^{k - 1}\big(1 - \pi^i(\boldsymbol{c_j}) \big)$ denotes the probability of the sample going through the residuals for all the previous iterations from $1$ through $k-1$ (\ie $\prod_{i=1} ^{k - 1}\big(1 - \pi^i(\boldsymbol{c_j}) \big)$\big) times the probability of going through the interpretable model at iteration $k$ \big(\ie $\pi^k(\boldsymbol{c_j})$\big). 
 At iteration $k$, $\pi^1, \dots \pi^{k - 1}$ are not trainable. 
 


\begin{table*}[t]
\caption{MoIE does not hurt the performance of the original BB. We 
provide AUROC and accuracy for medical imaging (\eg HAM10000, ISIC, and Effusion) and vision (\eg CUB-200 and Awa2) datasets, respectively, over 5 random seeds. For MoIE, we also
report the ``Coverage''. We only report the performance of the convolutional CEM~\cite{zarlenga2022concept}, leaving the construction of VIT-based CEM as future work. As HAM10000 and ISIC have no concept annotation, interpretable-by-design models can not be constructed. 
}
\fontsize{4.2pt}{0.28cm}\selectfont
\label{tab:performance}
\begin{center}
\begin{tabular}{p{27em} p{10em} p{9em} p{9em} p{9em} p{9em} p{9em} p{9em}}
\toprule 
        \textbf{MODEL} & \multicolumn{7}{c}{\textbf{DATASET}} \\
       & CUB-200 (RESNET101) & CUB-200 (VIT) & AWA2 (RESNET101) & AWA2 (VIT) & HAM10000 & SIIM-ISIC & EFFUSION  \\
\midrule 
    BLACKBOX & 0.88 & 0.92 & 0.89 & 0.99 & 0.96 & 0.85 & 0.91\\
\midrule
    \textbf{INTERPRETABLE-BY-DESIGN} \\
    CEM~\cite{zarlenga2022concept} & 0.77 $\pm$ 0.002 & - & 0.88 $\pm$ 0.005 & - & NA & NA & 0.76 $\pm$ 0.002\\
    CBM (Sequential)~\cite{koh2020concept} & 0.65 $\pm$ 0.003 & 0.86 $\pm$ 0.002 & 0.88 $\pm$ 0.003 & 0.94 $\pm$ 0.002  & NA & NA  
    & 0.79 $\pm$ 0.005 \\ 
    CBM + E-LEN~\cite{koh2020concept, barbiero2022entropy} & 0.71 $\pm$ 0.003 & 0.88 $\pm$ 0.002 & 0.86 $\pm$ 0.003 & 0.93 $\pm$ 0.002 & NA & NA & 
    0.79 $\pm$ 0.002  \\
\midrule
     \textbf{POSTHOC} \\
     PCBM~\cite{yuksekgonul2022post} & 0.76 $\pm$ 0.001  & 0.85 $\pm$ 0.002 & 0.82 $\pm$ 0.002 & 0.94 $\pm$ 0.001 &
     0.93 $\pm$	0.001 & 0.71 $\pm$	0.012 & 0.81 $\pm$	0.017\\
     PCBM-h~\cite{yuksekgonul2022post} & 0.85 $\pm$ 0.001  & 0.91 $\pm$ 0.001 & 0.87 $\pm$ 0.002 & 0.98 $\pm$ 0.001 &
     0.95 $\pm$	0.001 & 0.79 $\pm$	0.056 & 0.87 $\pm$	0.072\\
     PCBM + E-LEN~\cite{yuksekgonul2022post, barbiero2022entropy} &  0.80 $\pm$ 0.003 & 0.89 $\pm$ 0.002 & 0.85 $\pm$ 0.002 & 0.96 $\pm$ 0.001 & 
     0.94 $\pm$	0.021 &  0.73 $\pm$	0.011 & 0.81 $\pm$	0.014\\
     PCBM-h + E-LEN~\cite{yuksekgonul2022post, barbiero2022entropy} &  0.85 $\pm$ 0.003 & 0.91 $\pm$ 0.002 & 0.88 $\pm$ 0.002 & 0.98 $\pm$ 0.002 & 
     0.95 $\pm$	0.032 &  0.82 $\pm$	0.056 & 0.87 $\pm$	0.032\\
\midrule
     \textbf{OURS} \\
     MoIE (COVERAGE) &\textbf{0.86 $\pm$ 0.001 (0.9)} &\textbf{0.91 $\pm$ 0.001 (0.95)} &
     \textbf{0.87 $\pm$ 0.002 (0.91)} & \textbf{0.97 $\pm$ 0.004 (0.94)} & \textbf{0.95 $\pm$	0.001 (0.9)}
     & \textbf{0.84 $\pm$ 0.001 (0.94)} & \textbf{0.87 $\pm$	0.001 (0.98)}\\
     MoIE + RESIDUAL & \textbf{0.84 $\pm$ 0.001} & \textbf{0.90 $\pm$ 0.001} & \textbf{0.86 $\pm$ 0.002} & \textbf{0.94 $\pm$ 0.004}
     & \textbf{0.92 $\pm$	0.00} & \textbf{0.82 $\pm$	0.01} & \textbf{0.86 $\pm$	0.00} \\
\bottomrule
\end{tabular}
\end{center}
\end{table*}

\begin{table}[h]
\caption{Performance of various shortcut elimination methods on Waterbirds dataset.}
\fontsize{5.2pt}{0.30cm}\selectfont
\label{tab:shortcut}
\vskip 0.1in
\begin{center}
\begin{tabular}{l|c|c}
\toprule
Method & Avg Acc. & Worst Acc.  \\
\midrule
ERM~\cite{wah2011caltech}  & 97.0 $\pm$ 0.2\% & 63.7 $\pm$ 1.9\% \\
ERM+aug~\cite{wah2011caltech}  & 87.4 $\pm$ 0.5\% &  76.4 $\pm$ 2.0\% \\
UW~\cite{xian2018zero} & 96.3.0 $\pm$ 0.3\% & 76.2 $\pm$ 1.4\%\\
IRM~\cite{arjovsky2020invariant} & 87.5 $\pm$ 0.7\% & 75.6 $\pm$ 3.1\%\\
IB-IRM~\cite{ahuja2022invariance}    & 88.5 $\pm$ 0.9\% & 76.5 $\pm$ 1.2\%\\
V-REx~\cite{krueger2021outofdistribution}    & 88.0 $\pm$ 1.4\% & 73.6 $\pm$ 0.2\%\\
CORAL~\cite{sun2016deep}    & 90.3 $\pm$ 1.1\% & 79.8 $\pm$ 1.8\%\\
Fish~\cite{shi2021gradient}    & 85.6 $\pm$ 0.4\% & 64.0 $\pm$ 0.3\%\\
GroupDRO~\cite{sagawa2019dro}    & 91.8 $\pm$ 0.3\% & 90.6 $\pm$ 1.1\%\\
JTT~\cite{liu2021just}    & 93.3 $\pm$ 0.3\% & 86.7 $\pm$ 1.5\%\\
DM-ADA~\cite{Xu-2020}    & 76.4 $\pm$ 0.3\% & 53.0 $\pm$ 1.3\%\\
LISA~\cite{pmlr-v162-yao22b}    & 91.8 $\pm$ 0.3\% & 88.5 $\pm$ 0.8\%\\
\midrule
BB w MDN (\textbf{ours})    & \textbf{95.01 $\pm$ 0.5\%} & \textbf{94.4 $\pm$ 0.5\%}\\
MoIE from BB w MDN (\textbf{ours}) (COVERAGE)   & 91.0 $\pm$ 0.5\% (0.91) & 93.7 $\pm$ 0.4\% (0.87)\\
MoIE+R from BB w MDN (\textbf{ours})   & 90.2 $\pm$ 0.5\% & 92.1 $\pm$ 0.4\%\\
\bottomrule
\end{tabular}
\end{center}
\vskip -0.1in
\end{table}

\textbf{The Residuals:}
The last step is to \emph{repeat} with the residual $r^k$, as $\displaystyle r^k(\boldsymbol{x_j},\boldsymbol{c_j}) = f^{k - 1}(\boldsymbol{x_j}) - g^k(\boldsymbol{c_j})$.
We fix $\Phi$ and optimize the following loss to update $h^k$ to specialize on those samples not covered by $g^k$, effectively creating a new BB $f^k$ for the next iteration $(k+1)$:
\begin{equation}
\label{equ: residual}
\mathcal{L}_f^k(\boldsymbol{x_j}, \boldsymbol{c_j}) = \underbrace{\mystrut{2.6ex}\ell\big(r^k(\boldsymbol{x_j}, \boldsymbol{c_j}), f^k(\boldsymbol{x_j})\big)}_{\substack{\text{trainable component} \\ \text{for iteration $k$}}} \underbrace{\mystrut{2.6ex}\prod_{i=1} ^{k}\big(1 - \pi^i(\boldsymbol{c_j})\big)}_{\substack{\text{non-trainable component} \\ \text{for iteration $k$}}} 
\end{equation}


We refer to all the experts as the Mixture of Interpretable Experts (MoIE). We denote the experts, including the final residual, as MoIE+R. Each expert in MoIE constructs sample-specific FOLs using the optimization strategy in SelectiveNet~\cite{geifman2019selectivenet}.


\subsection{Applying to mitigate shortcuts}
~\cref{alg:shortcut} illustrates a 3-step procedure to eliminate shortcuts.
 The BB, trained on a dataset with shortcuts, latches on the spurious concepts to classify the labels. \textbf{Detection.} The FOLs from the biased BB-derived MoIE, capture the spurious concepts. \textbf{Elimination.} Assuming the spurious concepts as metadata, we minimize the effect of shortcuts from the representation of the BB using MDN~\cite{lu2021metadata} layers between two successive layers of the convolutional backbone to fine-tune the biased BB. MDN is a regression-based normalization technique to mitigate metadata effects and improve model robustness. \textbf{Verfitication.} Finally, we distill the MoIE from the new robust BB and generate the FOLs. The FOLs validate if the BB still uses spurious concepts for prediction.

%% file: Sections/experiments_v1.tex
We perform experiments to show that MoIE does not compromise the accuracy of the original BB across various datasets and architectures and eliminates shortcuts using the Waterbirds dataset \cite{sagawa2019dro}. As a stopping criterion, we repeat our method until MoIE covers at least 90\% of samples. Furthermore, we only include concepts as
input to $g$ if their validation accuracy or AUROC exceeds a certain threshold (in all of our experiments,
we fix 0.7 or 70\% as the threshold of validation auroc or accuracy). Refer to~\cref{tab:dataset} for the datasets and BBs' experimented with. For ResNets, Inception, and DenseNet121, we flatten the feature maps from the last convolutional block to extract the concepts. For VITs, we use the image embeddings from the transformer encoder to perform the same. We use SIIM-ISIC as a real-world transfer learning setting, with the BB trained on HAM10000 and evaluated on a subset of the SIIM-ISIC Melanoma Classification dataset~\cite{yuksekgonul2022post}.~\cref{app:dataset} and~\cref{app:g} expand on the datasets and hyperparameters. Furthermore, we utilize E-LEN, \ie a Logic Explainable Network~\cite{ciravegna2023logic} implemented with an Entropy Layer as first layer~\cite{barbiero2022entropy} as the interpretable symbolic model $g$ to construct FOL explanations of a given prediction. With ResNet50 as the BB for shortcut detection, we use MDN layers between convolution blocks.

\begin{figure}[h]
\begin{center}
\centerline{\includegraphics[width=\columnwidth]{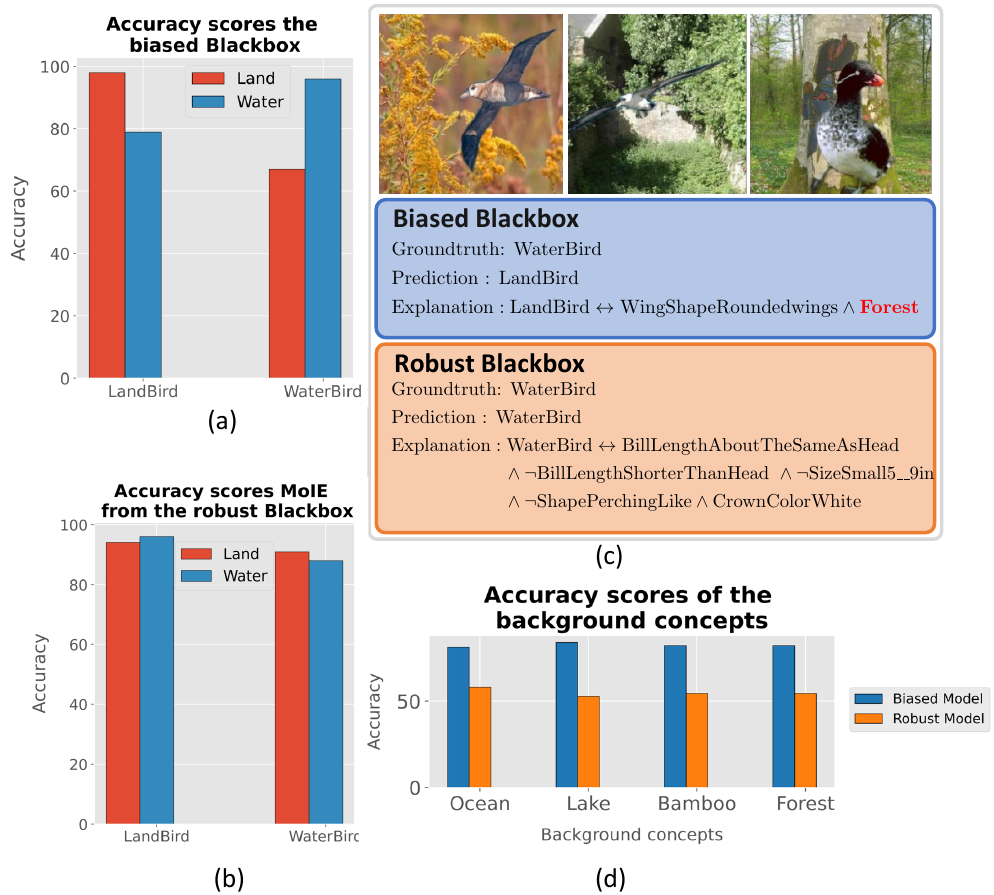}}
\caption{
MoIE fixes shortcuts. Performance of \textbf{(a)} the biased BB and 
\textbf{(b)} final MoIE extracted from the robust BB. 
\textbf{(c)} Examples of samples (\textbf{top-row}) and FOLs extracted from the biased (\textbf{middle-row}) and robust BB (\textbf{bottom-row}). 
\textbf{(d)} Accuracies of the spurious concepts extracted from the biased vs. the robust BB.
}
\label{fig:shortcut}
\end{center}
\vskip -0.2in
\end{figure}

\textbf{Baselines:}
To show the efficacy of our method compared to other concept-based models, we compare our methods to two concept-based baselines -- 1) interpretable-by-design and 2) posthoc. They consist of two parts: a) a concept predictor $\Phi: \mathcal{X} \rightarrow \mathcal{C}$, predicting concepts from images; and b) a label predictor $g: \mathcal{C} \rightarrow \mathcal{Y}$, predicting labels from the concepts. The end-to-end CEMs and sequential CBMs serve as interpretable-by-design baselines. Similarly, PCBM and PCBM-h serve as post hoc baselines. Convolution-based $\Phi$ includes all layers till the last convolution block. VIT-based $\Phi$ consists of the transformer encoder block. The standard CBM and PCBM models do not show how the concepts are composed to make the label prediction. So, we create CBM + E-LEN, PCBM + E-LEN, and PCBM-h + E-LEN by using the identical $g$ of MOIE (shown in~\cref{app:g}), as a replacement for the standard classifiers of CBM and 
PCBM. We train the $\Phi$ and $g$ in these new baselines to sequentially generate FOLs~\cite{barbiero2022entropy}. Due to the unavailability of concept annotations, we extract the concepts from the Derm7pt dataset~\cite{kawahara2018seven} using the pre-trained embeddings of the BB~\cite{yuksekgonul2022post} for HAM10000. Thus, we do not have interpretable-by-design baselines for HAM10000 and ISIC.

For shortcut-based methods, we compare our method with Empirical Risk Minimization (ERM) with and without data augmentations; Up- weighting (UW), which weights the instances of minority groups; Invariant Learning algorithms: IRM~\cite{arjovsky2020invariant}, IB-IRM~\cite{ahuja2022invariance}; Domain generalization/adaptation methods: V-REx~\cite{krueger2021outofdistribution}, CORAL~\cite{sun2016deep}, and Fish~\cite{shi2021gradient}; Instance reweighting methods: GroupDRO~\cite{sagawa2020distributionally}, JTT~\cite{liu2021just}; Data augmentation methods: DM-ADA~\cite{Xu-2020}, LISA~\cite{pmlr-v162-yao22b}.

%% file: Sections/quant_v1.tex
As MoIE uses multiple experts covering different subsets of data compared to the single one by the baselines, MoIE outperforms the baselines for most of the datasets, shown in~\cref{tab:performance}. Awa2 comprises rich concept annotation for zero-shot learning, resulting in better performance for the interpretable-by-design baselines.~\cref{app:qual} illustrates that  MoIE captures a diverse set of concepts qualitatively.~\cref{app:harder} shows that later iterations of MoIE cover the ``harder'' examples.

%% file: Sections/shortcut_v1.tex
\cref{tab:shortcut} demonstrates the efficacy of MoIE in eliminating the shortcuts than the other shortcut removal method by achieving high worst-case accuracy. First, the BB's accuracy differs for land-based versus aquatic subsets of the bird species, as shown in~\cref{fig:shortcut}a. The Waterbird on the water is more accurate than on land (96\%  vs. 67\% in the red bar). In the interpretable ``Detection'' stage, the FOLs from the biased BB-derived MoIE detect the spurious background concept \textit{forest} for a waterbird, misclassified as a landbird in~\cref{fig:shortcut}c (\emph{top row}). In the ``Elimination'' stage, the fine-tuned BB with MDN layers removes the specific background from the BB's representation ($\Phi$). Next, we train $t$, using $\Phi$ of the finetuned BB, and compare the accuracy of the spurious concepts with the biased BB in~\cref{fig:shortcut}d. The validation accuracy of all the spurious concepts retrieved from the finetuned BB falls well short of the predefined threshold of 70\% compared to the biased BB. Finally, we distill the MoIE from the robust BB.~\cref{fig:shortcut}b illustrates similar accuracies of MoIE for Waterbirds on water vs. Waterbirds on land (89\% - 91\%). As the shortcut concepts are removed successfully, MoIE outperforms the other shortcut removal methods in worst group accuracy in~\cref{tab:shortcut} (the last 2 rows). In the interpretable ``Verification'' stage, the FOL from the robust BB does not include any background concepts (~\ref{fig:shortcut}c, bottom row).

%% file: Sections/conclusion_v1.tex
This paper proposes a novel method to iteratively extract a mixture of interpretable models from a flexible BB to eliminate shortcuts. We aim to leverage MoIE to eliminate shortcuts with varying complexities in the future. 

%% file: Appendix/dataset_v1.tex
\paragraph{CUB-200}
The Caltech-UCSD Birds-200-2011 (\cite{wah2011caltech}) is a fine-grained classification dataset comprising 11788 images and 312 noisy visual concepts. The aim is to classify the correct bird species from 200 possible classes. We adopted the strategy discussed in~\cite{barbiero2022entropy} to extract 108 denoised visual concepts. Also, we utilize training/validation splits shared in \cite{barbiero2022entropy}. Finally, we use the state-of-the-art classification models Resnet-101 (\cite{he2016deep}) and Vision-Transformer (VIT) (\cite{wang2021feature}) as the blackboxes $f^0$. 


\paragraph{Animals with attributes2 (Awa2)}
 AwA2 dataset \cite{xian2018zero} consists of 37322 images of a total of 50 animal classes with 85 numeric attributes. We use the state-of-the-art classification models Resnet-101 (\cite{he2016deep}) and Vision-Transformer (VIT) (\cite{wang2021feature}) as the blackboxes $f^0$.

\paragraph{HAM10000}
HAM10000 (\cite{tschandl2018ham10000}) is a classification dataset aiming to classify a skin lesion as benign or malignant. Following \cite{daneshjou2021disparities}, we use Inception \cite{szegedy2015going} model, trained on this dataset as the blackbox $f^0$. We follow the strategy in \cite{lucieri2020interpretability} to extract the eight concepts from the Derm7pt (\cite{kawahara2018seven}) dataset.

\paragraph{SIIM-ISIC}
To test a real-world transfer learning use case, we evaluate the
model trained on HAM10000 on a subset of the SIIM-ISIC\cite{rotemberg2021patient}) Melanoma Classification dataset. We use
the same concepts described in the HAM10000 dataset.

\paragraph{MIMIC-CXR} We use  220,763 frontal images from the MIMIC-CXR dataset \cite{12_johnsonmimic} aiming to classify effusion. We obtain the anatomical and observation concepts from the RadGraph annotations in RadGraph’s inference dataset (\cite{10_jain2021radgraph}), automatically generated by DYGIE++ (\cite{23_wadden-etal-2019-entity}). We use the test-train-validation splits from \cite{yu2022anatomy} and Densenet121 \cite{huang2017densely} as the blackbox $f^0$. 

\paragraph{Waterbirds}
~\cite{sagawa2019dro} creates the Waterbirds dataset by using forest and bamboo as the spurious land concepts of the Places dataset for landbirds of the CUB-200 dataset. We do the same by using oceans and lakes as the spurious water concepts for waterbirds. We utilize ResNet50 as the Blackbox $f^0$ to identify each bird as a Waterbird or a Landbird.

%% file: Appendix/g_v1.tex
\cref{tab:bb_config} demonstrates different settings to train the Blackbox of CUB-200, Awa2 and MIMIC-CXR respectively. For the VIT-based backbone, we used the same hyperparameter setting used in the state-of-the-art Vit-B\_16 variant in \cite{wang2021feature}. To train $t$, we flatten the feature maps from the last convolutional block of $\Phi$ using ``Adaptive average pooling'' for CUB-200 and Awa2 datasets. For MIMIC-CXR and HAM10000, we flatten out the feature maps from the last convolutional block. For VIT-based backbones, we take the first block of representation from the encoder of VIT.  For HAM10000, we use the same Blackbox in \cite{yuksekgonul2022post}.~\cref{tab:g_config_cub_200},~\cref{tab:g_config_awa2},~\cref{tab:g_config_ham10k},~\cref{tab:g_config_mimic_cxr} enumerate all the different settings to train the interpretable experts for CUB-200, Awa2, HAM, and MIMIC-CXR respectively. All the residuals in different iterations follow the same settings as their blackbox counterparts.

\begin{table}[h]
\caption{Hyperparameter setting of different convolution-based Blackboxes used by CUB-200, Awa2 and MIMIC-CXR}
\label{tab:bb_config}
\begin{center}
\begin{tabular}{l c c c}
\toprule 
     {\textbf{Setting}} & {\textbf{CUB-200}} & {\textbf{Awa2}} & 
    {\textbf{MIMIC-CXR}}\\
\midrule 
       Backbone              & ResNet-101 & ResNet-101 & DenseNet-121  \\
       Pretrained on ImageNet      & True &True & True \\
       Image size            & 448 & 224 & 512 \\
       Learning rate         & 0.001 & 0.001 & 0.01 \\
       Optimization          & SGD & Adam & SGD \\
       Weight-decay      & 0.00001 & 0 & 0.0001 \\
       Epcohs             & 95 & 90 & 50 \\
       Layers used as $\Phi$ &  \makecell{till 4$^{th}$ ResNet \\Block} &  \makecell{till 4$^{th}$ ResNet \\Block} &  \makecell{till 4$^{th}$ DenseNet \\Block} \\
       Flattening type for the input to $t$    &  \makecell{Adaptive average \\pooling} &  \makecell{Adaptive average \\pooling} & Flatten \\
\bottomrule
\end{tabular}
\end{center}
\end{table}

\begin{table}[h]
\caption{Hyperparameter setting of interpretable experts ($g$) trained on ResNet-101 (top) and VIT (bottom) blackboxes for CUB-200 dataset}
\label{tab:g_config_cub_200}
\begin{center}
\begin{tabular}{l|c|c|c|c|c|c}
\toprule 
    \thead{\textbf{Settings based on dataset}} & \thead{\textbf{Expert1}} & \thead{\textbf{Expert2}} 
    & \thead{\textbf{Expert3}} & \thead{\textbf{Expert4}} & \thead{\textbf{Expert5}} & \thead{\textbf{Expert6}}\\
\midrule 
        CUB-200 (ResNet-101)              &    &   &  & &  & \\
       \quad + Batch size              & 16 & 16 & 16 & 16 & 16 & 16   \\
        
       \quad + Coverage ($\tau$)  & 0.2 & 0.2 & 0.2 & 0.2 & 0.2 & 0.2 \\
       
       \quad + Learning rate & 0.01 & 0.01 & 0.01 & 0.01 & 0.01 & 0.01 \\
       
       \quad + $\lambda_{lens}$ & 0.0001 & 0.0001 & 0.0001 & 0.0001 & 0.0001 & 0.0001 \\
    
       \quad +$\alpha_{KD}$ & 0.9 & 0.9 & 0.9 & 0.9 & 0.9 & 0.9 \\
       \quad + $T_{KD}$ & 10 & 10 & 10 & 10 &10 & 10 \\
       \quad +hidden neurons & 10 & 10 & 10 & 10 &10 & 10 \\
       \quad +$\lambda_s$ & 32 & 32 & 32 & 32 & 32 & 32 \\
       \quad + $T_{lens}$ & 0.7 & 0.7 & 0.7 & 0.7 & 0.7 & 0.7 \\
\midrule 
        CUB-200 (VIT)            &    &   &  & &  & \\
       \quad + Batch size              & 16 & 16 & 16 & 16 & 16 & 16   \\
        
       \quad + Coverage ($\tau$)  & 0.2 & 0.2 & 0.2 & 0.2 & 0.2 & 0.2 \\
       
       \quad + Learning rate & 0.01 & 0.01 & 0.01 & 0.01 & 0.01 & 0.01 \\

       \quad + $\lambda_{lens}$ & 0.0001 & 0.0001 & 0.0001 & 0.0001 & 0.0001 & 0.0001 \\
    
       \quad +$\alpha_{KD}$ & 0.99 & 0.99 & 0.99 & 0.99 & 0.99 & 0.99 \\
       \quad + $T_{KD}$ & 10 & 10 & 10 & 10 &10 & 10 \\
       \quad +hidden neurons & 10 & 10 & 10 & 10 &10 & 10 \\
       \quad +$\lambda_s$ & 32 & 32 & 32 & 32 & 32 & 32 \\
       \quad +$T_{lens}$ & 6.0 & 6.0 & 6.0 & 6.0 & 6.0 & 6.0 \\
\bottomrule
\end{tabular}
\end{center}
\end{table}

\begin{table}[h]
\caption{Hyperparameter setting of interpretable experts ($g$) trained on ResNet-101 (top) and VIT (bottom) blackboxes for Awa2 dataset}
\label{tab:g_config_awa2}
\begin{center}
\begin{tabular}{l|c|c|c|c|c|c}
\toprule 
    \thead{\textbf{Settings based on dataset}} & \thead{\textbf{Expert1}} & \thead{\textbf{Expert2}} 
    & \thead{\textbf{Expert3}} & \thead{\textbf{Expert4}} & \thead{\textbf{Expert5}} & \thead{\textbf{Expert6}}\\
\midrule 
        Awa2 (ResNet-101)              &    &   &  & &  & \\
       \quad + Batch size              & 30 & 30 & 30 & 30 & - & -   \\
        
       \quad + Coverage ($\tau$)  & 0.4 & 0.35 & 0.35 & 0.25 & - & - \\
       
       \quad + Learning rate & 0.001 & 0.001 & 0.001 & 0.001 & - & - \\
       
       \quad + $\lambda_{lens}$ & 0.0001 & 0.0001 & 0.0001 & 0.0001 & - & - \\
    
       \quad +$\alpha_{KD}$ & 0.9 & 0.9 & 0.9 & 0.9 & - & - \\
       \quad + $T_{KD}$ & 10 & 10 & 10 & 10 & - & - \\
       \quad +hidden neurons & 10 & 10 & 10 & 10 & - & - \\
       \quad +$\lambda_s$ & 32 & 32 & 32 & 32 & - & - \\
       \quad + $T_{lens}$ & 0.7 & 0.7 & 0.7 & 0.7 & - & - \\
\midrule 
        Awa2 (VIT)            &    &   &  & &  & \\
       \quad + Batch size              & 30 & 30 & 30 & 30 & 30 & 30   \\
        
       \quad + Coverage ($\tau$)  & 0.2 & 0.2 & 0.2 & 0.2 & 0.2 & 0.2 \\
       
       \quad + Learning rate & 0.01 & 0.01 & 0.01 & 0.01 & 0.01 & 0.01 \\

       \quad + $\lambda_{lens}$ & 0.0001 & 0.0001 & 0.0001 & 0.0001 & 0.0001 & 0.0001 \\
    
       \quad +$\alpha_{KD}$ & 0.99 & 0.99 & 0.99 & 0.99 & 0.99 & 0.99 \\
       \quad + $T_{KD}$ & 10 & 10 & 10 & 10 &10 & 10 \\
       \quad +hidden neurons & 10 & 10 & 10 & 10 &10 & 10 \\
       \quad +$\lambda_s$ & 32 & 32 & 32 & 32 & 32 & 32 \\
       \quad + $T_{lens}$ & 6.0 & 6.0 & 6.0 & 6.0 & 6.0 & 6.0 \\
\bottomrule
\end{tabular}
\end{center}
\end{table}

\begin{table}[h]
\caption{Hyperparameter setting of interpretable experts ($g$) for the dataset HAM10000}
\label{tab:g_config_ham10k}
\begin{center}
\begin{tabular}{l|c|c|c|c|c|c}
\toprule 
    \thead{\textbf{Settings based on dataset}} & \thead{\textbf{Expert1}} & \thead{\textbf{Expert2}} 
    & \thead{\textbf{Expert3}} & \thead{\textbf{Expert4}} & \thead{\textbf{Expert5}}
    & {\textbf{Expert6}}\\
\midrule 
        HAM10000 (Inception-V3)              &    &   &  & & &  \\
       \quad + Batch size              & 32 & 32 & 32 & 32 & 32&  32   \\
        
       \quad + Coverage ($\tau$)  & 0.4 & 0.2 & 0.2 & 0.2 & 0.1&  0.1\\
       
       \quad + Learning rate & 0.01 & 0.01 & 0.01 & 0.01 & 0.01& 0.01 \\
       
       \quad + $\lambda_{lens}$ & 0.0001 & 0.0001 & 0.0001 & 0.0001 & 0.0001 &  0.0001\\
    
       \quad +$\alpha_{KD}$ & 0.9 & 0.9 & 0.9 & 0.9 & 0.9& 0.9\\
       \quad + $T_{KD}$ & 10 & 10 & 10 & 10 & 10& 10 \\
       \quad +hidden neurons & 10 & 10 & 10 & 10 & 10& 10\\
       \quad +$\lambda_s$ & 64 & 64 & 64 & 64 & 64& 64  \\
       \quad + $T_{lens}$ & 0.7 & 0.7 & 0.7 & 0.7 & 0.7& 0.7 \\
\bottomrule
\end{tabular}
\end{center}
\end{table}

\begin{table}[h]
\caption{Hyperparameter setting of interpretable experts ($g$) for the dataset MIMIC-CXR}
\label{tab:g_config_mimic_cxr}
\begin{center}
\begin{tabular}{l|c|c|c}
\toprule 
    \thead{\textbf{Settings based on dataset}} & \thead{\textbf{Expert1}} & \thead{\textbf{Expert2}} 
    & \thead{\textbf{Expert3}} \\
\midrule 
        Effusion-MIMIC-CXR (DenseNet-121)              &    &   &     \\
       \quad + Batch size              & 1028 & 1028 & 1028     \\
        
       \quad + Coverage ($\tau$)  & 0.6 & 0.2 & 0.15   \\
       
       \quad + Learning rate & 0.01 & 0.01 & 0.01 \\
       
       \quad + $\lambda_{lens}$ & 0.0001 & 0.0001 & 0.0001  \\
    
       \quad +$\alpha_{KD}$ & 0.99 & 0.99 & 0.99  \\
       \quad + $T_{KD}$ & 20 & 20 & 20   \\
       \quad +hidden neurons & 20, 20 & 20, 20 & 20, 20  \\
       \quad +$\lambda_s$ & 96 & 128 & 256   \\
       \quad +$T_{lens}$ & 7.6 & 7.6 & 7.6 \\
\bottomrule
\end{tabular}
\end{center}
\end{table}

%% file: Appendix/qual_v1.tex
\begin{figure*}[t]
\vskip 0.1in
\begin{center}
\centerline{\includegraphics[width=\linewidth]{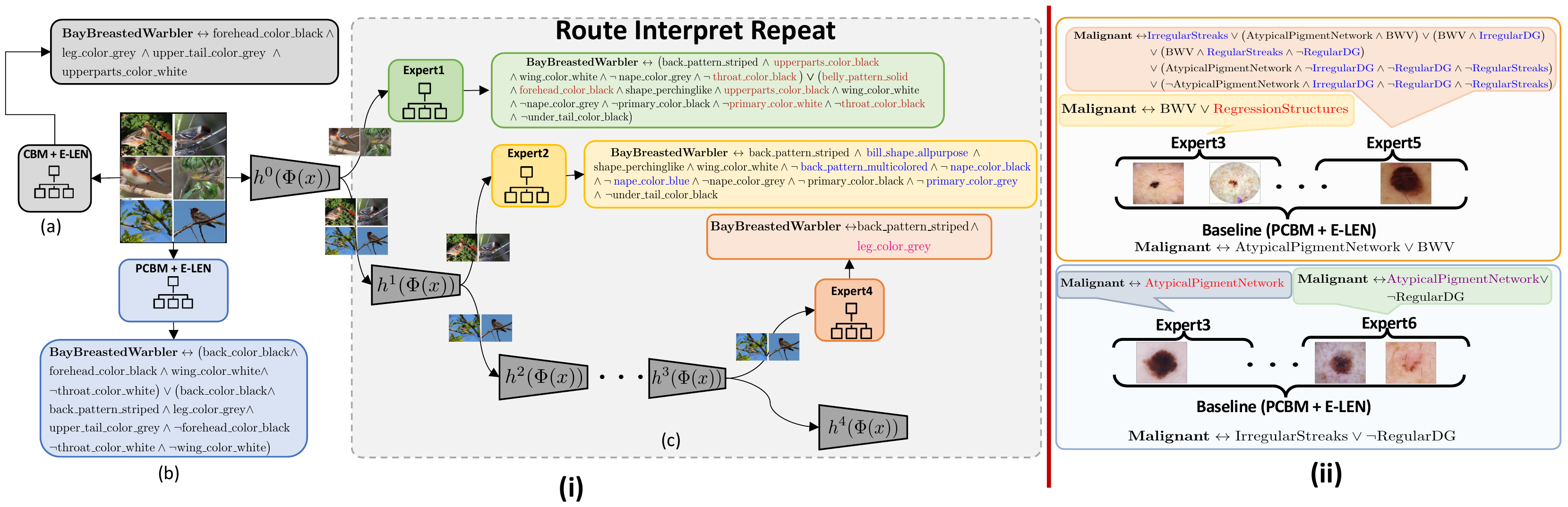}}
\caption{MoIE identifies diverse concepts for specific subsets of a class, unlike the generic ones by the baselines. \textbf{(i)} We construct the FOL explanations of the samples of, ``Bay breasted warbler'' in the CUB-200 dataset for VIT-based \textbf{(a)} CBM + E-LEN as an \emph{interpretable-by-design} baseline, \textbf{(b)} PCBM + E-LEN as a \emph{posthoc} baseline, \textbf{(c)} experts in MoIE at inference. We highlight the unique concepts for experts 1,2, and 3 in~\emph{red},~\emph{blue}, and~\emph{magenta}, respectively. \textbf{(ii)} Comparison of FOL explanations by MoIE with the PCBM + E-LEN baselines for HAM10000 (\textbf{top}) and ISIC (\textbf{down})  to classify Malignant lesion. We highlight unique concepts for experts 3, 5, and 6 in \emph{red}, \emph{blue}, and \emph{violet}, respectively. For brevity, we combine the local FOLs for each expert for the samples covered by them, shown in the figure.}
\label{fig:local_ex_cub}
\end{center}
\vskip -0.1in
\end{figure*}

\begin{figure*}[h]
\centering
\includegraphics[width=1.0\textwidth]
{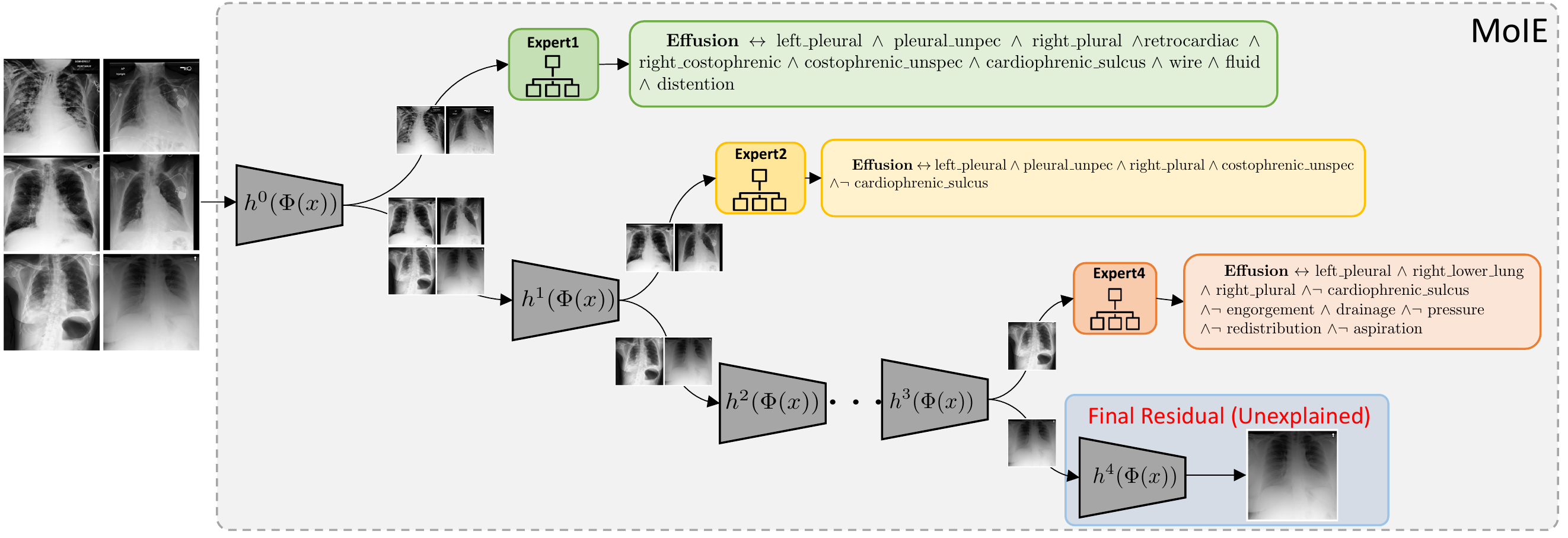}
\caption{Construction logical explanations of the samples of ``Effusion'' in the MIMIC-CXR dataset for various experts in MoIE at inference. The final residual covers the unexplained sample, which is ``harder'' to explain (indicated in \emph{red}).}
\label{fig:mimic_concept_explanation}
\end{figure*}


\begin{figure*}[h]
\centering
\includegraphics[width=\columnwidth]{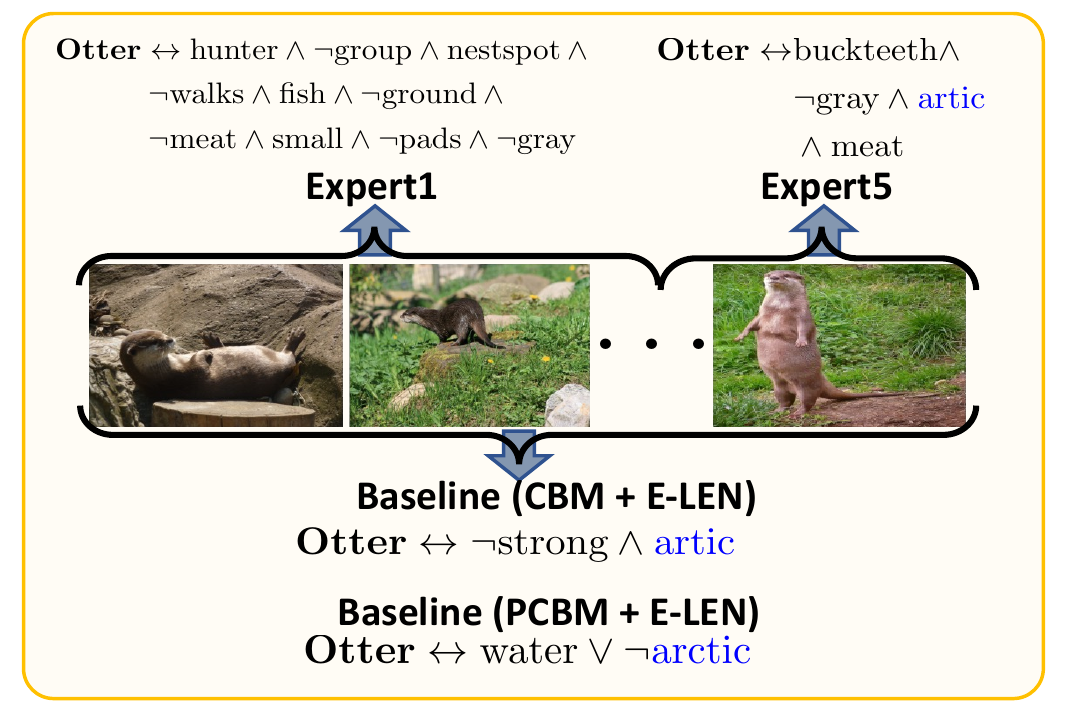}
\vspace{-10pt}
\caption{Flexibility of FOL explanations by VIT-derived MoIE  MoIE and the CBM + E-LEN and PCBM + E-LEN baselines for Awa2 dataset to classify ``Otter'' at inference. Both the baseline's FOL constitutes identical concepts to distinguish all the samples. However, expert1 classifies ``Otter'' with \textit{hunter}, \textit{group} \etc as the identifying concept for the instances covered by it. Similarly expert5 classifies ``Otter'' using \textit{buckteeth}, \textit{gray} \etc. Note that, \textit{meat} and \textit{gray}  are shared between the two experts. We highlight the shared concepts (\textit{artic}) between the experts and the baselines as blue.}
\label{fig:local_awa2_otter}
\vspace{-2.5pt}
\end{figure*}

\begin{figure*}[h]
\centering
\includegraphics[width=20cm,
  height=20cm,
  keepaspectratio]{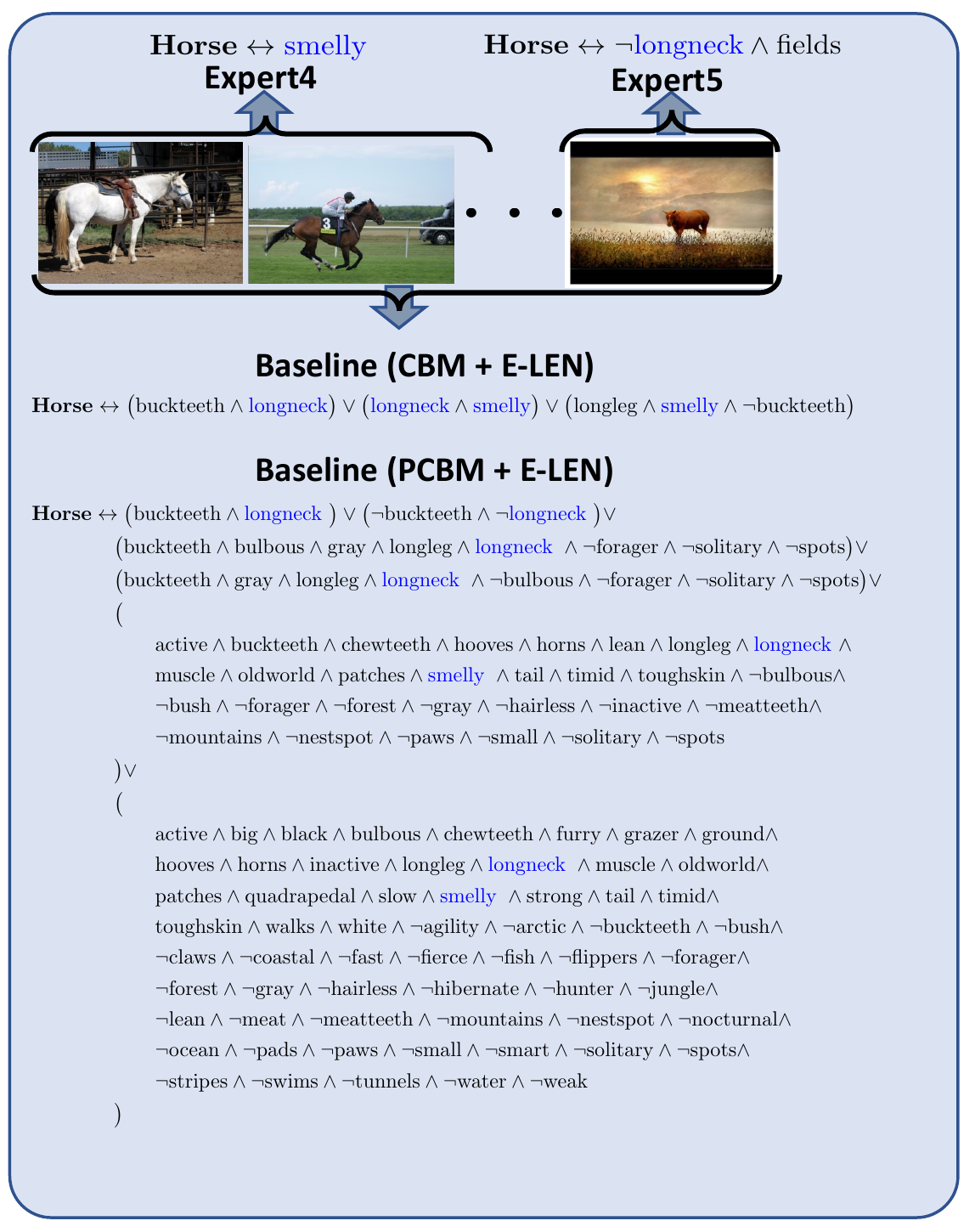}
\vspace{-10pt}
\caption{Flexibility of FOL explanations by VIT-derived MoIE  MoIE and the CBM + E-LEN and PCBM + E-LEN baselines for Awa2 dataset to classify ``Horse'' at inference. Both the baseline's FOL constitutes identical concepts to distinguish all the samples. However, expert4 classifies ``Horse'' with \textit{smelly} as the identifying concept for the instances covered by it. Similarly, expert5 classifies the same ``Horse'' using \textit{longneck} and \textit{fields}. We highlight the shared concepts between the experts and the baselines as blue.}
\label{fig:local_awa2_horse}
\vspace{-2.5pt}
\end{figure*}

\textbf{Heterogenity of Explanations:}
At each iteration of MoIE, the blackbox \big($h^k(\Phi(.)$\big) splits into an interpretable expert ($g^k$) and a residual ($r^k$).~\cref{fig:local_ex_cub}i shows this mechanism for VIT-based MoIE and compares the FOLs with CBM + E-LEN and PCBM + E-LEN baselines to classify ``Bay Breasted Warbler'' of CUB-200.
The experts of different iterations specialize in specific instances of ``Bay Breasted Warbler''. Thus, each expert's FOL comprises its instance-specific concepts of the same class. For example, the concept, \emph{leg\_color\_grey} is unique to expert4, but \emph{belly\_pattern\_solid} and \emph{back\_pattern\_multicolored} are unique to experts 1 and 2, respectively, to classify the instances of ``Bay Breasted Warbler'' in the ~\cref{fig:local_ex_cub}(i)-c. 
Unlike MoIE, the baselines employ a single interpretable model $g$, resulting in a generic FOL with identical concepts for all the samples of ``Bay Breasted Warbler'' (\cref{fig:local_ex_cub}i(a-b)). Thus the baselines fail to capture the heterogeneity of explanations. Due to space constraints, we combine the local FOLs of different samples.

\cref{fig:local_ex_cub}ii shows such diverse local instance-specific explanations for HAM10000 (\emph{top}) and ISIC (\emph{bottom}). In~\cref{fig:local_ex_cub}ii-(top), the baseline-FOL consists of concepts such as \emph{AtypicalPigmentNetwork} and \emph{BlueWhitishVeil (BWV)} to classify ``Malignancy'' for all the instances for HAM10000. However, expert~3 relies on \emph{RegressionStructures} along with \emph{BWV} to classify the same for the samples it covers. At the same time, expert~5 utilizes several other concepts \eg \emph{IrregularStreaks}, \emph{Irregular dots and globules (IrregularDG)} \etc \text{ }Due to space constraints,~\cref{fig:local_awa2_otter} demonstrates similar results for the Awa2 dataset.

%% file: Appendix/harder_v1.tex
\begin{figure*}[ht]
\vskip 0.2in
\begin{center}
\centerline{\includegraphics[width=\linewidth]{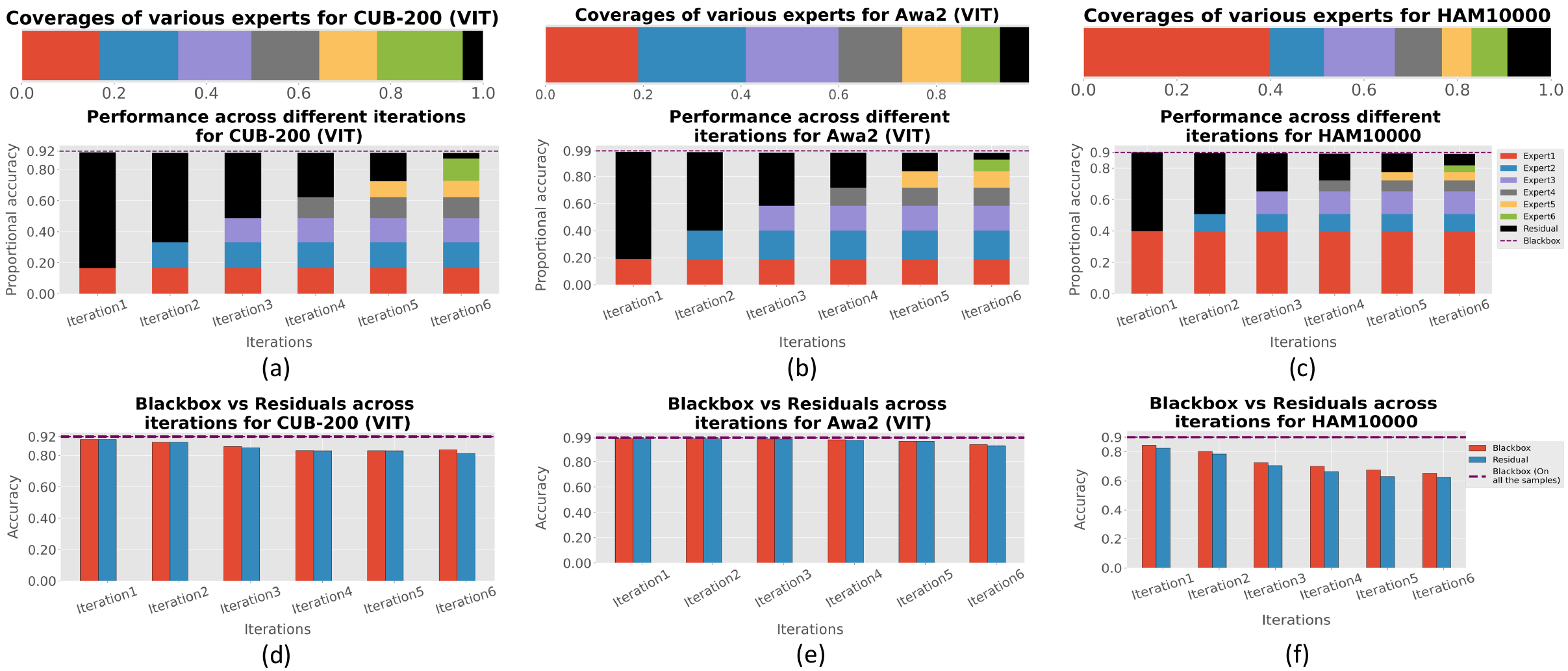}}
\caption{The performance of experts and residuals across iterations. 
\textbf{(a-c)} Coverage and proportional accuracy of the experts and residuals. 
\textbf{(d-f)} We route the samples covered by the residuals across iterations to the initial Blackbox $f^0$ and compare the accuracy of $f^0$ (red bar) with the residual (blue bar). Figures \textbf{d-f} show the progressive decline in performance of the residuals across iterations as they cover the samples in the increasing order of ``hardness''. We observe the similar abysmal performance of the initial blackbox $f^0$ for these samples.
}
\label{fig:expert_performance_cv_vit}
\end{center}
\vskip -0.2in
\end{figure*}

\begin{figure}[h]
\centering
\includegraphics[width=1\linewidth]{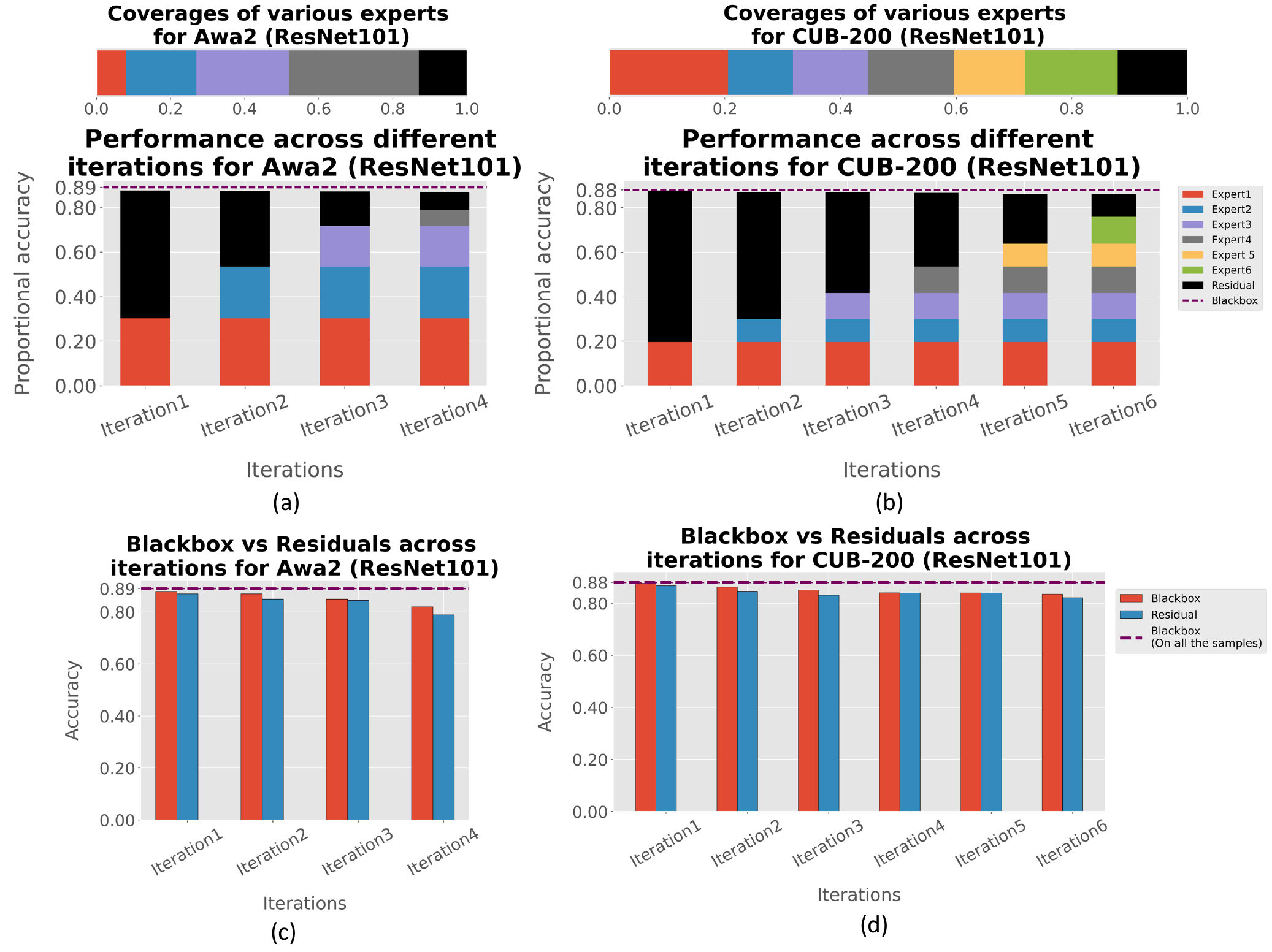}
\caption{The performances of experts and residuals across iterations for ResNet derived MoIE for CUB-200 and Awa2. 
\textbf{(a-b)} Coverage and proportional accuracy of the experts and residuals. 
\textbf{(c-d)} We route the samples covered by the residuals across iterations to the initial Blackbox $f^0$ and compare the accuracy of $f^0$ (red bar) with the residual (blue bar).}
\label{fig:expert_performance_cv_resnet}
\end{figure}

\cref{fig:expert_performance_cv_vit} (a-c) display the proportional accuracy of the experts and the residuals of our method per iteration. The proportional accuracy of each model (experts and/or residuals) is defined as the accuracy of that model times its coverage. Recall that the model's coverage is the empirical mean of the samples selected by the selector. 
\cref{fig:expert_performance_cv_vit}a show that the experts and residual cumulatively achieve an accuracy $\sim$ 0.92 for the CUB-200 dataset in iteration 1, with more contribution from the residual (black bar) than the expert1 (blue bar). Later iterations cumulatively increase and worsen the performance of the experts and corresponding residuals, respectively. The final iteration carves out the entire interpretable portion from the Blackbox $f^0$ via all the experts, resulting in their more significant contribution to the cumulative performance. The residual of the last iteration covers the ``hardest'' samples, achieving low accuracy. Tracing these samples back to the original Blackbox $f^0$, it also classifies these samples poorly (\cref{fig:expert_performance_cv_vit}{(d-f)}).
As shown in the coverage plot, this experiment reinforces~\cref{fig:Schematic}, where the flow through the experts gradually becomes thicker compared to the narrower flow of the residual with every iteration. 
\cref{fig:expert_performance_cv_resnet} shows the coverage (top row), performances (bottom row) of each expert and residual across iterations of - (a) ResNet101-derived Awa2 and (b) ResNet101-derived CUB-200 respectively.